# A Survey on Patients Privacy Protection with Stganography and Visual Encryption

Hussein K. Alzubaidy[1], Dhiah Al-Shammary[1], *Mohammed Hamzah Abed[1]

[1]College of Computer Science and Information TechnologyUniversity of Al-Qadisiyah Dewaniyah Iraq

1*mohammed.abed@qu.edu.iq

* Corresponding Author

**Abstract.** In this survey, thirty models for steganography and visual encryption methods have been discussed to provide patients privacy protection. Image steganography techniques are classified into two domains: Spatial Domain and Frequency Domain. Spatial hiding models are divided into three groups: Least Significant Bit (LSB), Most Significant Bit (MSB) and other spatial hiding while Frequency models are divided into two groups: Discrete cosine transform (DCT) and Discrete Wavelet Transform (DWT). Moreover, some models have been used to encrypt medical images. Technically, PSNR (Peak-Signal-to-Noise-Ratio) have been computed and compared to investigate and evaluate the model's performance. As a result of this survey analysis, Frequency Hiding models for DCT (Discrete Cosine Transform) have shown the best Average PSNR with 70.66 dB, compared to other approaches.

## 1 Introduction

The rapid growth of the Internet and information technology has led to exposure to patient data. It has become easy to attack and tamper medical images [23]. To avoid data loss and mishandling, several steganography techniques are used for privacy protection. Steganography can be classified into two domains: Spatial and Frequency. In the spatial domain, secret messages are directly hidden in the cover image. Pixel values are manipulated to investigate the desired improvement. On the other hand, the frequency domain is based on the conversion of the pixel from Time-domain to the Frequency domain. Cryptography is the process of encrypting messages in a way that hackers cannot understand the content. One of the traditional methods for cryptography is Advanced Encryption Standard (AES) which is an asymmetric cypher. Furthermore, visual encryption is another form of cryptography dedicated to visual images to be encrypted or scrambled.

## 1.1 Motivation

Lately, cases of privacy violation for personal data is growing rapidly and tremendously with the increased usage of the internet [14]. Moreover, sudden attacks on organizations and modifications that can be occurred to secret information, the main challenge is assuring security for Electronic Patient Records (EPRs) and preservation of data integrity from unauthorized access in telemedicine applications and healthcare services. Concealing confidential information is important at securing communication. Therefore, it is necessary to discover effective and powerful methods to hide such information through using the steganography technique that allows hiding the content of data during transmission from eyes intruders.

## 1.2 Survey Strategy and Evaluation

This survey presented thirty Publications for steganography techniques described and classified into six groups. Three groups represent spatial hiding models for medical images: Least significant bit (LSB), Most Significant Bit (MSB), and other Spatial hiding models. Furthermore, two groups for frequency hiding models: Discrete Cosine Transforms (DCT) and Discrete Wavelet Transform (DWT). Finally, visual encryption models are discussed in the last group. To evaluate the success of the proposed models, the best PSNR results are collected and analyzed in tables and charts. Frequency hiding models have achieved the highest average PSNR compared to other approaches.

## 1.3 Paper Organization

The rest of this paper is organized as follow: Section 2 describes spatial hiding models for medical images that include three groups: Least significant bit (LSB), Most significant bit (MSB) and other spatial hiding. Frequency Hiding models for medical images is presented in section 3 that includes two groups: Discrete Cosine Transform(DCT) and Discrete Wavelet Transform(DWT). Visual encryption models for medical images are illustrated in section 4. Analysis and evaluation are presented in section 5. Finally, conclusions are presented in section 6.

## 2 Spatial Hiding Models

Secret messages are directly hidden in the cover image. Pixel values are manipulated to investigate the desired improvement. This section has discussed Least significant bit (LSB), the Most significant bit(MSB) and other spatial hiding.

### 2.1 Least significant bit (LSB)

The Least Significant Bit (LSB) is the most common steganography method based on manipulating the least significant bits in image pixels. G. F. Siddiqui et al [2] has discussed the Improvement of data masking ability for medical stego-images and patient records. Image Region Decomposition(IRD) has been proposed to divide (MRI) images into three regions based on intensity. To evaluate their proposed solution, PSNR, SSIM, MSE values have been computed and compared with other techniques. The authors have used ten medical images for evaluation. furthermore, another five standard images like Lena and cameraman have been used for evaluation to compare with other models. The best-achieved result in this research is 49.27 PSNR. Technically, few medical image samples have been evaluated that would not prove the persistency of the proposed method. C. Lee et al [3] has discussed the concealment of sensitive information for medical records without an increase in data size. This research has proposed a data hiding method by using a magic cube. Moreover, they have proposed the Least Significant Bit (LSB) method as a partial hiding technique. then the differences are computed and contrasted with the coordinate position in the 3D magic cube. To evaluate their proposed solution, PSNR, MSE, ER have been computed and compared to investigate their results. This paper has used nine standard images like Lena and Baboon in addition to a few medical images for evaluation. The best-achieved result in this research is 2.25 an embedding capacity with 44 average PSNR. Few image samples have been evaluated and the model performance has not been investigated. Unauthorized disclosure of medical images for electronic patient's records (EPR) and hiding have been addressed by M. Ulutas et al [ 7]. This paper has proposed a (k, n) Shamir's secret sharing scheme to hide (EPR) and prevent unauthorized access of medical information. Technically, PSNR, MSE and NOC (Number of characters) have been computed and compared to investigate their results. Eleven medical images have been used for evaluation. The best-achieved result in this research is 46.3702 PSNR. Although high performance has been achieved by the proposed method, the Reconstructed image contains removed regions and might be susceptible to damage during transmission.

S. Karakus et al [ 14] have discussed the increasing amount of data to be hidden and stego-image to reduce deterioration medical image quality. A new optimization method has been proposed based on optimum pixel similarity. Technically, PSNR, MSE, RMSE, SSIM, UQI have been computed and compared to investigate their results. This paper has used twenty images for both MR, CT and OT in the evaluation process. The best-achieved result in this research is 66.5374 PSNR for 1000 characters' data is hidden in $512^x512$ images. On the other hand, 10,000 characters of data hiding cannot be hidden in $256^x256$ images without compression techniques. Therefore, the optimization process is hard.

Another study by R. Bhardwaj et al [15] that has discussed provision saving for telemedicine applications during the communication process and exchange of patient information. A dual image separable block has been proposed based on a reversible data hiding algorithm. Technically, PSNR, MSE, embedding capacity, BER has been computed and compared to investigate their results. This paper has used twenty general images like Lena and Boat and medical images for evaluation. The best-achieved result in this research is 54.10 PSNR. Finally, the proposed method has achieved potential PSNR, However, it is still weak for resistance of various attacks on images .in addition, the proposed method has not been proven its efficiency with processing time.

R. F. Mansour et al [16], Have addressed sharing electronic patient information in telemedicine systems and improving the security of patients services overcloud. RDH (Reversible Data Hiding) has been proposed based on Lagrange's interpolation polynomial, secret sharing and bit substitution. To evaluate their proposed solution, PSNR, payload, entropy, standard deviation, and cross-correlation have been computed and compared to investigate their results. This paper has used Sixteen general and medical images for evaluation. The best-achieved result in this research is 52.38 PSNR. Several image samples have been tested and the proposed method performance is suffering from high distortion and lack of capacity.

Hiding electronic patient records (EPR) in medical images by using LSB (Least significant bit) is proposed by N. A. Loan et al [ 20]. Reversible data hiding schemes have been proposed based on Pixel Repetition Method(PRM) and edge detection. Technically, PSNR, MSE, SSIM, BER have been computed and compared to investigate their results. This paper has used twenty general images and ten medical images for evaluation. The best-achieved result in this research is 58.832 PSNR. On the other hand, few image samples are used that does not confirm the high efficiency.

Hiding medical images for patients to preserve the integrity and security of medical information has been discussed by C. Kim et al [21]. A stenographic data hiding scheme (HDH) has been proposed. This proposed method is a combination between Hamming code and LSB with an optimal pixel adjustment process algorithm. Technically, PSNR, MSE has been computed and compared to investigate their results. This paper has used ten medical images for evaluation. The best-achieved result in this research is 52.44 PSNR. Although The proposed method has achieved potential PSNR, ten image samples are still few to prove their performance.

D.Wang et al [22], Have addressed avoiding malicious attack and tampering during data transmission for EHR (Electronic health records) for security privacy patients. A high steganography capacity spatial domain data hiding scheme has been proposed to hide information by using four LSB (Least Significant Bit). Technically, PSNR, Weighted PSNR (w PSNR), MSE, SSIM have been computed and compared to investigate their results. This paper has used ten medical images for evaluation. The best-achieved result in this research is 66.08 PSNR. On the other hand, Image quality deterioration is not obvious, Average loss of PSNR is 0.45. Therefore, the proposed method is not guaranteed to improve concealment capacity.

Another research by G. Gao et al [ 24] that have discussed using new methods for hiding secret data and medical images with lossless and low distortion. Reversible Data Hiding with an Automatic contrast enhancement algorithm (RDHACEM) have been proposed for hiding medical images. This proposed method separates medical images into Region of interest (ROI) and non-ROI. To evaluate their proposed solution, PSNR, SSIM, RCE (Relative Contrast Error), Relative Mean Brightness Error (RMBE) and Mean Opinion Score (MOS) have been computed and compared to investigate their results. This paper has used six medical images with twelve marked images for each image and four algorithms (RHCRDH, ACERDH, RDHMBP and RDHACEM) for evaluation. The best-achieved result in this research is 26.5136 PSNR for Brain -1 image. Although several images have been tested, image robustness and data security protections are not clarified.

A. AbdelRaouf et al [ 30] have addressed protecting secret messages from unauthorized access through using the steganography technique. A new data hiding model has been proposed based on the human visual system(HVS) and using Least Significant Bits (LSB). Technically, PSNR, MSE, SSIM have been computed and compared to investigate their results. This paper has used six images as the dataset for evaluation. The best-achieved result in this research is 84.48 as an average PSNR. Finally, few images have been tested, it is not enough to get an accurate evaluation to have.

## 2.2    Most Significant Bit (MSB)

Most Significant Bit hiding methods are considered to be sensitive as they may cause a high error in comparison with the original image. L. Yang et al [ 4], Has discussed the weakness of traditional LSB and how to increase secret information protection, Most Significant Bit (MSB) has been proposed to increase hiding capacity and reduce stego- image distortion to improve its quality and reduce bit error rate(BER) when distortion occurs. Moreover, they have applied BOW (Bag of Words), this model extracts visible words(VW) to represent confidential information. To evaluate their proposed solution, PSNR, MSE, SSIM, QI have been computed and compared to investigate their results. The proposed model has achieved potential PSNR; However, it is not clear how to resist distortion. The best-achieved result in this research for hiding {2601} bits of secret information with PSNR near to infinity. although high performance has been investigated in hiding information, it's still prone to error bit rate and sudden attacks.

A. A. Abd El-Latif et al [ 25] have discussed concealing quantum secret images inside a cover image. Two new and efficient information hiding approaches have been proposed based on MSQb and LSQb. To evaluate their proposed solution, PSNR, MSE has been computed and compared to investigate their results. This paper has used six as Cover and secret images and nine as Carrier and watermark images for evaluation. The best-achieved result in this research is 48.1897 PSNR for Carrier and watermark images and 46.7901 PSNR for cover and secret images. On the other hand, several images have been tested, the proposed method is not potentially investigated on known data.

## 2.3    Other Spatial hiding models

Other spatial hiding methods are depicted in this section. M. Li et al [1] has addressed tracing illegal disclosure of medical images in a multicast communication environment and how to define a set of requirements, in addition to the challenges to enhance patient's privacy. This research has proposed multicast fingerprinting based on IA-W watermarking. Moreover, invertible watermarking is proposed for integrity protection. To evaluate their proposed solution, PSNR, QI, MSME have been computed and compared to investigate their results This paper has been used thirty-one images with five modalities for evaluation. The best-achieved results in this research are 86.34 PSNR with a compressed ratio of 10:1. Although best performance has been achieved, still tracing is harder because of the number of comparisons.

J.A. Kawa et al [5], Has discussed Protecting electronic patient records(EPR) from unauthorized access in the medical image. This research has proposed Optimal Pixel Repetition (OPR) to secure (EPR) inside a medical image. Moreover, the reversible framework is proposed in three phases. To evaluate their proposed solution, PSNR, NCC, SSIM, BER, NAE have been computed and compared to investigate their results. Nine images have been used for evaluation that was obtained from several databases. The best-achieved results in this research are 42 average PSNR. The proposed model requires significant extraction time when hiding a large amount of data.

Maximizing and improving hiding capacity for medical images has been discussed by O. M. Al-Qershi et al [ 27]. Two reversible data hiding schemes have been proposed based on Difference Expansion(DE). Furthermore, this proposed method combine Chiang with Tian and AL attar. To evaluate their proposed solution, PSNR, SSIM has been computed and compared. This paper has used sixteen DICOM images for evaluation. The best-achieved result in this research is 95.59 PSNR for Chiang and 82.34 PSNR for (Tian and Chiang) method with (embedding capacity =0.1). On the other hand, a small number of image samples have been tested. Moreover, there is no clear security metric as most metrics are for image quality.

S.A. Parah et al [29], Have discussed the provision high degree of security for EHR in the healthcare system by hiding images for smart city applications. This paper has proposed high capacity, secure and computationally efficient Electronic Health Record (EHR) hiding for medical images in Internet of Things (IoT) based healthcare technique. Moreover, the proposed technique has used Pixel Repetition Method(PRM) and modular arithmetic. Technically, PSNR, MSE, NAE, NCC have been computed and compared to investigate their results. This paper has used twelve images includes six general images and six medical images. In addition, twenty UCID images for evaluation. The best-achieved result in this research is 45.42 PSNR. On the other hand, the proposed technique has not shown clear efficiency.

## 3  Frequency Hiding models

Based on the manipulation of the orthogonal transform of the image instead of the image itself. This section has addressed Discrete Cosine Transforms (DCT), Discrete Wavelet Transform (DWT).

### 3.1 Discrete Cosine Transforms (DCT)

Discrete Cosine Transform is a commonly used mathematical transformation the has been applied in several areas including frequency-based hiding approaches. N. Ayub et al [ 6], Has discussed hiding data in a highly secure position. This paper has proposed a novel technique to hide secret information along edges inside images. several filters have been used to detect edges and data is hiding by DCT. In order to evaluate their proposed solution, PSNR, MSE, SNR have been computed and compared to investigate their results. Ten BMP and jpeg images have been used for evaluation. The proposed model has resulted in the best PSNR of 84.9 with a canny filter for BMP images. furthermore, best-achieved PSNR for jpeg images is 99.8 with a canny filter. Although the proposed model has achieved potential PSNR, the resultant image size reduction might affect intruders.

X. Liao et al [9], Has discussed how to conceal medical JPEG images and provide safeguard for patient's medical information. A new medical JPEG image stenographic scheme has been proposed that relied on dependencies of inter-block coefficients. In order to evaluate their proposed solution, nzAC, CR, CEC, ECR have been computed and compared to investigate their results. This paper has used twelve images for evaluation. The best-achieved result in this research is less than 1% as incremental testing error ratio. The destruction in inter-block dependencies of DCT coefficients has caused potential weakness in performance.

Hiding a medical image into another medical image to maintain on electronic patient records from the leakage is discussed by A. Elhadad et al [13]. This paper has proposed a steganography technique based on DICOM medical images. Moreover, the proposed method included three main parts: preprocessing, data embedding based on DCT and an extraction process. In order to evaluate their proposed solution, PSNR, MSE, SSIM, UQI, Correlation coefficient (R) have been computed and compared to investigate their results. This paper has used twenty patients for the evaluation process. The best-achieved result in this research is 85.39 PSNR. Finally, several image samples have been evaluated. However, a compression ratio is still low (varies between 0.4770 to 0.4988) of files size.

S. Arunkumar et al [ 26], Have addressed securing secret medical images and sensitive information in healthcare from unauthorized use during transmission. A robust image steganographic method has been proposed based on combined Redundant Integer Wavelet Transform (RIWT), Discrete Cosine Transforms (DCT), Singular Value Decomposition (SVD) and the logistic chaotic map. Technically, PSNR, IF, NAE, MSE, MSSIM, NCC, have been computed and compared to investigate their results. This paper has used eight images for evaluation and four for comparison. The best-achieved result in this research is 50.18 PSNR and the average PSNR of the proposed method for four images have been selected Lena, Airplane, Pepper and Baboon is 49.77 compared to other methods. Finally, few image samples have been used and that still is not enough for testing.

Another hiding method proposed by C.N. Yang et al [28] has addressed enhancement hiding capacity and maintaining quality for JPEG images. RDHS (Adaptive real-time reversible data hiding scheme) has been proposed based on discrete cosine transformation (DCT) coefficients that have been rearranged in a zigzag manner. In order to evaluate their proposed solution, PSNR, Embedding Capacity(EC) have been computed and compared. This paper has used six stego-images like Zelda and Baboon for evaluation. The best-achieved result in this research is 47.27 PSNR. On the other hand, the proposed method requires one more pass before the data is embedded or decoded. Therefore, it is hard and needs more time to execution. Moreover, few images have been tested and not enough to evaluate their performance.

**3.2 Discrete Wavelet Transform (DWT)**

Discrete Wavelet Transform is another common mathematical transform that is being used widely for steganography models. U. Subramaniyam et al [10], Has discussed secure data communication and hiding secret messages from hacking. has been proposed to connect and separate images based on frequency segments. Technically, PSNR, MSE has been computed and compared to investigate their results. This paper has used three JPEG images for both cover and secret images in the evaluation process. The best-achieved result in this research is 53.771 PSNR. Finally, very few image samples have been tested.

H. mohan Pandey et al [11], Have discussed medical sensitive data safety during transmission and reception. Bitmask oriented genetic algorithm (BMOGA) has been proposed to reduce redundancy of medical tests data while transmission. Furthermore, they have proposed steganography as an additional hiding technique. PSNR, SC, SSIM, MSE, BER and correlation have been computed and compared to investigate their results. Five medical images have been used for evaluation. The best-achieved result in this research is 74.69 PSNR. Finally, the processing time is not clear. Therefore, it is significantly hard to decide model to apply in real world.

Another study by S. Borraa et al [ 17] has discussed the provision of security and integrity for color radiological images during transmission. A hybrid image hiding technique with high capacity has been proposed based on Fast Discrete Curvelet Transform (FDCuT) and Compressive Sensing (CS for hiding color secret images. Technically, PSNR, MSE, SSIM, MS-SSIM have been computed and compared to investigate their results. This paper has used five images like knee MRI and Liver US for evaluation. The best-achieved result in this research is 70.27 PSNR. Few medical images are not enough to assure efficiency.

**4 Visual Encryption models**

Visual encryption is one of the well-known technical concepts to protect patient privacy through visually encrypting patient visual records. S. Kaur et al [8], Has addressed encryption and decryption to conceal secret images. IHED (Image Hiding Encryption and Decryption) has been proposed to encrypt and decrypt images. Moreover, the encryption process is carried out using the Mid Frequency (MF). They have been specified by Mid Search African Buffalo Model (MSABM). Technically, PSNR, MSE, SSIM have been computed and compared to investigate their results. This paper has used five MRI images like Heart and Liver for evaluation. The best-achieved result in this research is 71.6677 PSNR. Few images have been evaluated and the model proposed is inefficient against different types of attacks.

M. Elhoseny et al [12], Have addressed Securing and hiding confidential patient data in the electronic healthcare system, in this paper, a hybrid encryption model has been proposed to encrypt secret patient's data using two algorithms are RSA and AES through a combination of 2D-DWT-1L or 2D-DWT-2L steganography technique with the proposed model. The encoded data is hidden inside the cover image, then embedded data is extracted and decoded to retrieve the original image. Technically, PSNR, MSE, SSIM, BER, SC, and correlation have been computed and compared to investigate their results. This paper has used five images for both color and gray images for the evaluation process. The best-achieved results in this research are 57.44 PSNR and 56.09 PSNR in both color images and gray images respectively. On the other hand, the processing time is not computed. Therefore, it is significantly hard to assess model efficiency in real scenario.

Mechanisms to analyze and enhance multilevel techniques against security threats has been discussed by P. Panwar et al [18]. These aims investigate more integrity and security of ensuring hiding capacity for confidential patient information into image. Moreover, the impact of potential attack and noise is clear on communication channels. A hybrid encryption model has been proposed based on Quantum chaos(QC) and RSA encryption. Furthermore, encrypted data is embedded into image using the Improved BPCS technique for getting stego-image. In order to evaluate their proposed Solution PSNR, MSE, BER, SSIM, SC, UIQI, MAE, JI, BC, IC, CC, have been computed and compared to investigate their results. This paper has used five images for evaluation. The best-achieved result in this research is 71.38 PSNR in absence of attacks. Finally, the proposed method has achieved a high level of security. However, it is not proven to have high efficiency with testing on five images.

B. Prasanalakshmi et al [ 19], Have discussed secure medical data for patients during it transmission in IOT. HECC (Hyper elliptic Curve Cryptography) have been proposed based on AES algorithm, Blowfish hybrid cryptography, Koblitz method. Technically, PSNR, MAE, SSIM, SC and correlation have been computed and compared to investigate their results. This paper has used eight sample images for evaluation. The best-achieved result in this research is 70 PSNR. Finally, many modern methods have been compared and tested. Furthermore, the processing time is not clarified.

Another study by S. Shastri et al [23] has addressed securing privacy secret data from leakage and unauthorized access using Encryption and steganography techniques. A dual image RDH algorithm has been proposed based on the Centre Folding Strategy (CFS) and the Shiftable Pixel Coordinate Selection Strategy. In order to evaluate their proposed solution, PSNR, MSE, SSIM, Embedding Rate have been computed and compared. This paper has used six images like Lena and Barbara for evaluation. The best-achieved result in this research is 50.66 PSNR. Finally, a number of bits might cause problems in overflow or underflow if the embedding process is low, therefore the quality image depends on the number of pixels. In addition, the execution time of the proposed method is slowest compared to other techniques.

## 5   Analysis and Evaluation

This paper has presented thirty models, twenty-five for steganography techniques and five for Visual Encryption models. The analysis process has relied on the extraction results of PSNR from all papers in order to evaluate model's efficiency. Table 1 ,Table 2 and Table 3 shows the best results PSNR for spatial hiding models for medical images distributed into three groups: (Least Significant Bit (LSB), Most Significant Bit (MSB) and Other Spatial hiding models). Furthermore, Table 4 illustrates the best resultant PSNR for frequency hiding models for medical image distributed into two groups: Discrete Cosine Transform (DCT), Discrete Wavelet Transform (DWT). Table 5 Explain Best resultant PSNR for Visual Encryption models. On the other hand, obtained results in Tables 1, 2,3,4 and 5 are represented visually in the bar chart as shown in figure 1.

**Table 1.** Best PSNR for spatial hiding models with Least significant bit (LSB) for medical images

| Category | Models | PSNR dB |
|---|---|---|

| Least significant bit (LSB) | Image Region Decomposition (IRD) based on intensity [2] | 49.27 |
|---|---|---|
| | Data hiding method by using magic cube and Least Significant Bit (LSB) [3] | 44 |
| | a (k, n) Shamir's secret sharing scheme[7] | 46.3702 |
| | New optimization method has been proposed based on optimum pixel similarity[ 14] | 66.5374 |
| | Dual image separable block based reversible data hiding algorithm[15] | 54.10 |
| | RDH (Reversible Data Hiding) based on Lagrange's interpolation polynomial, secret sharing and bit substitution[16] | 52.38 |
| | Reversible data hiding scheme based Pixel Repetition Method (PRM) and edge detection [20] | 58.832 |
| | A steganographic data hiding scheme (HDH) with OPAP optimal pixel adjustment process algorithm [21] | 52.44 |
| | A high steganography capacity spatial domain data hiding scheme [22] | 66.08 |
| | Reversible Data Hiding with Automatic contrast enhancement algorithm (RDHACEM) [24] | 26.5136 |
| | Two new and efficient information hiding approaches based on MSQb and LSQb [25] | 46.7901 |
| | A new data hiding model based human visual system(HVS) [30] | 84.48 |

**Table 2.** Best PSNR for spatial hiding models with Most significant bit (MSB) for medical images

| Category | Models | PSNR dB |
|---|---|---|
| Most significant bit (MSB) | Most Significant Bit (MSB) model [4] | PSNR near to infinity |
| | Two new and efficient information hiding approaches based on MSQb and LSQb [25] | 48.1897 |

**Table 3.** Best PSNR for Other spatial hiding models for medical images

| Category | Models | PSNR |
|---|---|---|

|  |  | dB |
|---|---|---|
| **Most significant bit (MSB)** | multicast fingerprinting based on IA- W watermarking [1] | 86.34 |
|  | Optimal Pixel Repetition (OPR) and The reversible framework [5] | 42 |
|  | Two reversible data hiding schemes based Difference Expansion(DE) [27] | 95.59 |
|  | high capacity, secure and computationally efficient Electronic Health Record (EHR) hiding with Pixel Repetition Method(PRM) and modular arithmetic [29] | 45.42 |

**Table 4.** Best PSNR for frequency with DCT and DWT for medical images

| Category | Models | PSNR dB |
|---|---|---|
| **Discrete Cosine Transforms (DCT)** | A novel technique to hide secret information along edges inside images [6] | 99.8538 |
|  | New medical JPEG image steganographic scheme based dependencies of inter-block Coefficients [9] | _ |
|  | steganography technique based on DICOM medical image [13] | 85.39 |
|  | A robust image steganographic method based combine Redundant Integer Wavelet Transform (RIWT), Discrete Cosine Transforms (DCT), Singular Value Decomposition (SVD) and the logistic chaotic map [26] | 50.18 |
|  | RDHS (Adaptive real-time reversible data hiding scheme) based Discrete Cosine Transformation (DCT) coefficients [28] | 47.27 |
| **Discrete wavelet transform (DWT)** | Discrete wavelet transform (DWT) method [10] | 53.771 |
|  | BMOG (Bit mask oriented genetic algorithm) [11] | 74.69 |
|  | A hybrid image hiding techniques with high capacity based Fast Discrete Curve let Transform (FDCuT) and Compressive Sensing (CS) [17] | 70.27 |

**Table 5.** Best PSNR for visual encryption models for medical images

| Category | Models | PSNR dB |
|---|---|---|
| **Visual Encryption models** | IHED (Image Hiding Encryption and Decryption) [8] | 71.6677 |
|  | A hybrid encryption model using two algorithms are RSA and AES [12] | 57.44 |
|  | A hybrid encryption model has based Quantum chaos(QC) and RSA encryption [18] | 71.38 |
|  | HECC (Hyper elliptic Curve Cryptography) based on AES algo- | 70 |

| | |
|---|---|
| rithm, Blowfish hybrid cryptography, Koblitz method [19] | |
| A dual image RDH algorithm based on the Centre Folding Strategy (CFS) and the Shiftable Pixel Coordinate Selection Strategy [23] | 50.66 |

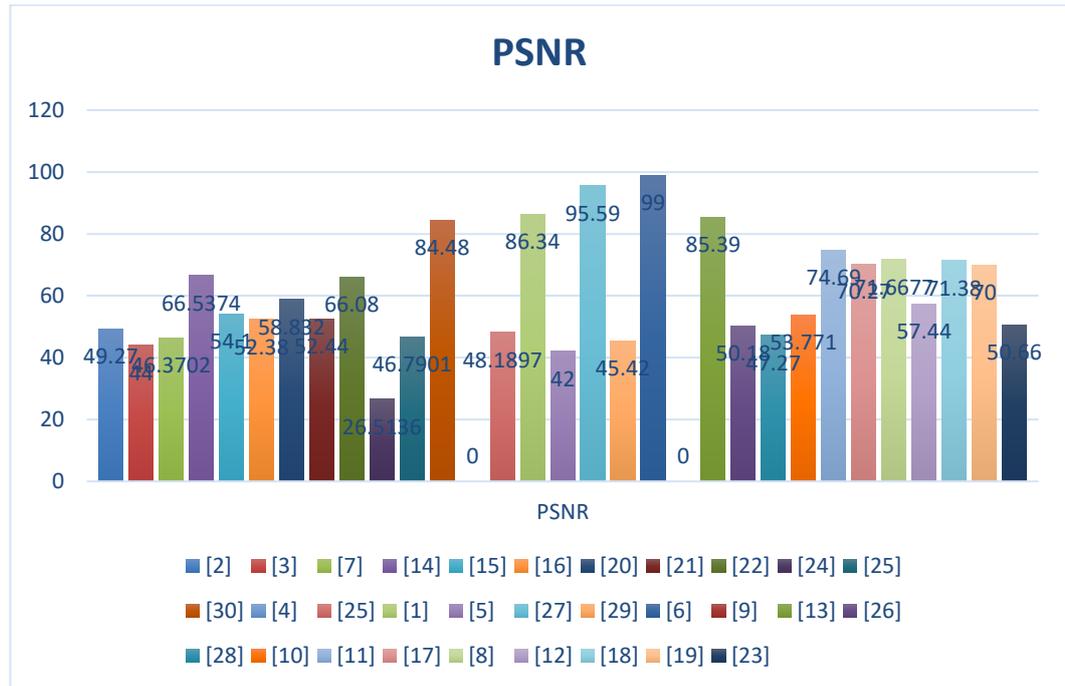

**Figure 1** PSNR values for different methods and techniques of steganography

## 6  Conclusion

In conclusion, this paper has presented thirty models divided into spatial and frequency domains for hiding the medical image for patients. Furthermore, visual encryption models for encrypting medical images. Hiding spatial models are divided into Least Significant Bit (LSB), Most significant Bit(MSB) and Other spatial hiding models.  Hiding frequency models are divided into Discrete cosine transform (DCT), Discrete wavelet transform (DWT). Technically, PSNR is extracted from the literature of the included models to evaluate models efficiency. Average PSNR is calculated for all groups and compared with each other. Average PSNR has showed outperformed hiding frequency models with 70.66 PSNR for Discrete Cosine Transforms (DCT) while average PSNR for hiding spatial models for a Least significant bit, Most Significant Bit and other Spatial hiding models are 53.9827, 48.1897 and 67.3375 respectively and average PSNR for visual encryption models is 64.22954.